\documentclass[sigconf,table,dvipsnames]{acmart}

\usepackage[utf8]{inputenc} % allow utf-8 input
\usepackage[T1]{fontenc}    % use 8-bit T1 fonts
\usepackage{hyperref}       % hyperlinks
\usepackage{url}
\hypersetup{
    colorlinks=true,
    linkcolor=red,
    citecolor=teal,
    urlcolor=cyan,
}
\usepackage{booktabs}       % professional-quality tables
\usepackage{amsfonts}       % blackboard math symbols
\usepackage{nicefrac}       % compact symbols for 1/2, etc.
\usepackage{microtype}      % microtypography
\usepackage{courier}
\usepackage{multirow}
\usepackage{lineno}
\usepackage{colortbl}
\usepackage{longtable}
\usepackage{listings}
\usepackage{amsmath}
\usepackage{bm}
\usepackage{graphicx}
\usepackage{tikz, tikz-cd}
\usetikzlibrary{positioning, arrows.meta}
\usepackage{mathtools}
\usepackage{subfig}
\usepackage{float}
\usepackage{wrapfig}
\usepackage{makecell}   % easy line breaks inside a cell
\usepackage{xcolor}
\usepackage{array}     % For fixed spacing
\usepackage{pifont}
\usepackage{newunicodechar}

\newcommand{\xmark}{\ding{55}}
\newcommand{\cmark}{\ding{51}}
\usepackage{subfig}
\usepackage{fancyvrb}
\usepackage{tcolorbox}
\usepackage{tabularx}
\usepackage{listings}
\lstdefinestyle{conceptlist}{
    basicstyle=\fontfamily{\sfdefault}\selectfont, % Use a sans-serif font, inherits size from context
    breaklines=true,            % Allow automatic line breaks
    breakatwhitespace=true,     % Break lines only at white space
    showstringspaces=false,     % Do not make spaces visible with special symbols
    frame=none,                 % No frame around the listing
    tabsize=2,                  % Set tab size (if you use tabs in your listing content)
    escapeinside={(*@}{@*)},    % Allows LaTeX commands between (*@ and @*)
    columns=flexible,           % Flexible column handling for better text flow
}

\definecolor{customgreen}{HTML}{01844F} % 3D9F3C
\definecolor{Green}{HTML}{3d9f3c}

\AtBeginDocument{%
  }

\setcopyright{acmlicensed}
\copyrightyear{2025}
\acmYear{2025}
\setcopyright{cc}
\setcctype{by-nc-sa}
\acmConference[MM '25]{Proceedings of the 33rd ACM International Conference on Multimedia}{October 27--31, 2025}{Dublin, Ireland}
\acmBooktitle{Proceedings of the 33rd ACM International Conference on Multimedia (MM '25), October 27--31, 2025, Dublin, Ireland}\acmDOI{10.1145/3746027.3758241}
\acmISBN{979-8-4007-2035-2/2025/10}

\begin{document}

\title{eSkinHealth: A Multimodal Dataset for \\ Neglected Tropical Skin Diseases}

% \author{Janet Wang}
% \email{swang47@tulane.edu}
% \affiliation{%

\author{Janet Wang}
\affiliation{%
  \institution{Tulane University}
  \city{New Orleans}
  \country{United States}}
\email{swang47@tulane.edu}

\author{Xin Hu}
\affiliation{%
  \institution{Tulane University}
  \city{New Orleans}
  \country{United States}}
\email{xhu13@tulane.edu}

\author{Yunbei Zhang}
\affiliation{%
  \institution{Tulane University}
  \city{New Orleans}
  \country{United States}}
\email{yzhang111@tulane.edu}

\author{Diabate Almamy}
\affiliation{%
  \institution{Université Alassane Ouattara}
  \city{Bouake}
  \country{Côte D'Ivoire}}
\email{docalmamy@yahoo.fr}

\author{Vagamon Bamba}
\affiliation{%
  \institution{Université de Bouaké}
  \city{Bouake}
  \country{Côte D'Ivoire}}
\email{bambavagamon@yahoo.com}

\author{Konan Amos Sébastien Koffi}
\affiliation{%
  \institution{Université de Bouaké}
  \city{Bouake}
  \country{Côte D'Ivoire}}
\email{askofi@hopecommission.org}

\author{Yao Koffi Aubin}
\affiliation{%
  \institution{Bakke Graduate University}
  \city{Dallas}
  \country{United States}}
\email{aubin@hopecommission.org}

\author{Zhengming Ding}
\affiliation{%
  \institution{Tulane University}
  \city{New Orleans}
  \country{United States}}
\email{zding1@tulane.edu}

\author{Jihun Hamm}
\affiliation{%
  \institution{Tulane University}
  \city{New Orleans}
  \country{United States}}
\email{jhamm3@tulane.edu}

\author{Rie R. Yotsu}
\authornote{Corresponding author.}
\affiliation{%
  \institution{Tulane University}
  \city{New Orleans}
  \country{United States}}
\email{ryotsu@tulane.edu}

\renewcommand{\shortauthors}{Wang et al.}
\renewcommand{\shorttitle}{eSkinHealth: A Multimodal Dataset for Neglected Tropical Skin Diseases}

\begin{abstract}
  Skin Neglected Tropical Diseases (NTDs) impose severe health and socioeconomic burdens in impoverished tropical communities. Yet, advancements in AI-driven diagnostic support are hindered by data scarcity, particularly for underrepresented populations and rare manifestations of NTDs. Existing dermatological datasets often lack the demographic and disease spectrum crucial for developing reliable recognition models of NTDs. To address this, we introduce \textbf{eSkinHealth}, a novel dermatological dataset collected on-site in Côte d’Ivoire and Ghana. Specifically, eSkinHealth contains 5,623 images from 1,639 cases and encompasses 47 skin diseases, focusing uniquely on skin NTDs and rare conditions among West African populations. We further propose an AI-expert collaboration paradigm to implement foundation language and segmentation models for efficient generation of multimodal annotations, under dermatologists' guidance. In addition to patient metadata and diagnosis labels, eSkinHealth also includes semantic lesion masks, instance-specific visual captions, and clinical concepts. Overall, our work provides a valuable new resource and a scalable annotation framework, aiming to catalyze the development of more equitable, accurate, and interpretable AI tools for global dermatology. Details about the code and dataset are available \href{https://github.com/janet-sw/eSkinHealth.git} {\texttt{\textcolor{blue}{\textbf{here}}}}.
  % The eSkinHealth dataset is accessible from \href{https://dataverse.harvard.edu/previewurl.xhtml?token=841dd5f1-5102-4d6c-a357-c3d2483817c3}{\textcolor{magenta}{here}}, and the codes from \href{https://github.com/janet-sw/SkinBench}{\textcolor{magenta}{here}}. 
  \textcolor{magenta}{Warning: This dataset may contain disturbing images of skin disease.}
\end{abstract}





\maketitle

\begin{table*}[htbp]
\centering
\caption{Comparison of public dermatology datasets. For data type, ``derm" is short for dermoscopic images.}
\resizebox{0.95\textwidth}{!}{%
\begin{tabular}{l|llrrcccc}
\toprule
\textbf{Geographic Origin} & \textbf{Dataset} & \textbf{Data Type} & \textbf{\#Sample} & \textbf{\#Class} & \textbf{Metadata} & \textbf{\#Concept} & \textbf{Caption} & \textbf{Mask} \\
\midrule
Unspecified & Derm7pt\cite{Kawahara2018-7pt} & derm\&clinical & 1,711 & 2 & \xmark & 7 & \xmark & \xmark \\
\midrule
\multirow{12}{*}{Europe/Oceania/South America }& HAM10000\cite{tschandl2018ham10000} & derm & 10,015 & 7 & \textcolor{magenta}{\cmark} & - & \xmark & \textcolor{magenta}{\cmark} \\
 & DermNet\cite{dermnet2023} & clinical & 19,500 & 23 & \xmark & - & \xmark & \xmark \\
 & ISIC 2016\cite{gutman2016skin} & derm & 1,271 & 2 & \xmark & - & \xmark & \textcolor{magenta}{\cmark} \\
 & ISIC 2017\cite{codella2018skin} & derm & 2,750 & 3 & \xmark & - & \xmark & \textcolor{magenta}{\cmark} \\
 & ISIC 2018\cite{codella2019skin, tschandl2018ham10000} & derm & 12,500 & 7 & \xmark & - & \xmark & \textcolor{magenta}{\cmark} \\
 & ISIC 2019\cite{hernandez2024bcn20000, codella2018skin, tschandl2018ham10000} & derm & 25,331 & 9 & \xmark & - & \xmark & \textcolor{magenta}{\cmark} \\
 & ISIC 2020\cite{rotemberg2021patient} & derm & 33,126 & 2 & \textcolor{magenta}{\cmark} & - & \xmark & \textcolor{magenta}{\cmark} \\
 & Fitzpatrick17k\cite{groh2021evaluating} & clinical & 16,577 & 114 & \xmark & - & \xmark & \xmark \\
 & PAD-UFES-20\cite{pacheco2020pad} & clinical & 2,298 & 6 & \textcolor{magenta}{\cmark} & - & \xmark & \xmark \\
 & Dermofit\cite{ballerini2013color} & clinical & 1,300 & 10 & \textcolor{magenta}{\cmark} & - & \xmark & \textcolor{magenta}{\cmark} \\
 & SkinCon\cite{daneshjou2022skincon} & clinical & 4,346 & 178 & \xmark & 48 & \xmark & \xmark \\
 & SkinCap\cite{zhou2024skincap} & clinical & 4,000 & 178 & \xmark & 48 & \textcolor{magenta}{\cmark} & \xmark \\
\midrule
\multirow{3}{*}{East Asia} & SD-198\cite{sun2016benchmark} & clinical & 6,584 & 198 & \xmark & - & \xmark & \xmark \\
 & XiangyaDerm\cite{xie2019xiangyaderm} & clinical & 107,565 & 541 & \textcolor{magenta}{\cmark} & - & \xmark & \xmark \\
 & Asan\cite{han2018classification} & clinical & 17,125 & 12 & \xmark & - & \xmark & \xmark \\
\midrule
\multirow{2}{*}{North America} & DDI\cite{daneshjou2022disparities} & clinical & 656 & 64 & \textcolor{magenta}{\cmark} & - & \xmark & \xmark \\
              & SCIN\cite{10.1001/jamanetworkopen.2024.46615} & clinical & 10,408 & - & \textcolor{magenta}{\cmark} & - & \xmark & \xmark \\

\midrule
\rowcolor{gray!20} West Africa & \textbf{eSkinHealth} & clinical & 5,623 & 47 & \textcolor{magenta}{\cmark} & 69 & \textcolor{magenta}{\cmark} & \textcolor{magenta}{\cmark} \\
\bottomrule
\end{tabular}%
}
\label{datasets_comparision}
\end{table*}

\section{Introduction}
Neglected Tropical Diseases (NTDs) is a diverse group of skin conditions prevalent in impoverished tropical communities, inflicting severe and multifaceted burdens. According to the WHO's 2024 Global Report on NTDs, these diseases result in significant mortality and morbidity, with approximately 120,000 deaths and 14.1 million disability-adjusted life years lost annually. Beyond the direct health toll, NTDs impose substantial economic costs on developing communities, equivalent to billions of U.S. dollars each year through direct health expenditures, lost productivity, and diminished socioeconomic and educational attainment. Furthermore, NTDs lead to profound social consequences, including disability, stigmatization, social exclusion, and discrimination, placing considerable financial strain on patients and their families. In the report, it emphasizes that early detection is critical to timely intervention and treatment of skin NTDs. However, limitations in medical personnel and diagnostic resources in these endemic regions create an urgent need for efficient testing tools to improve diagnostic accessibility \cite{malecela2021road, world2024global, yotsu2023deep}.

Recent advancements in computer vision have spurred the development of open-source skin disease datasets, enabling research into accurate and reliable models for diagnostic assistance \cite{Esteva2017DermatologistlevelCO, Liu2020-vx, Brinkerarticle, soenksen2021using, celebi2019dermoscopy, cai2023multimodal}. While training such models invariably requires large, diverse, and high-quality data, leading to significant collection efforts within the skin imaging analysis community, a critical gap still remains. Existing dermatology datasets, though valuable, are largely insufficient for advancing diagnostic models specifically for skin NTDs. Current datasets are typically collected in regions with different epidemiological profiles, often non-endemic to skin NTDs, and thus represent patient demographics, disease spectra, and symptom presentations that differ substantially from those characteristic of skin NTDs in affected communities.

To address these pressing gaps, we introduce \textbf{eSkinHealth}, a novel dermatological dataset collected on-site in Côte d’Ivoire and Ghana. This dataset is distinguished by its unique demographic focus, its emphasis on skin Neglected Tropical Diseases (NTDs), and its inclusion of rare skin conditions often absent from previous corpora. Specifically, eSkinHealth comprises 5,623 images from 1,639 cases, encompassing 47 distinct skin diseases, with patient metadata and confirmed diagnosis. Photographs were captured by local nurses or health workers trained in dataset acquisition. The diagnostic process for remotely assessed cases involved independent evaluation by two experienced dermatologists; in instances of disagreement, a third dermatologist was consulted to facilitate a consensus. Furthermore, a portion of cases received in-person diagnosis by a dermatologist during clinical visits, with some undergoing clinical tests to ensure diagnostic accuracy.

Leveraging recent advancements in foundation models, we investigated an AI-expert collaboration paradigm for generating multimodal annotations and semantic masks under dermatologists' guidance. For image captions and concepts, our process began with crafting condition-specific checklists derived from multiple credible sources, including the World Health Organization (WHO). The clinical validity of each checklist was then meticulously verified by board-certified dermatologists. Subsequently, we extracted a set of clinical concepts designed to discriminatively encode the pertinent features of each condition. These verified checklists and extracted concepts formed the foundation for guiding an advanced Multimodal Large Language Model (MLLM) to produce instance-specific captions and concepts for each image. Similarly, for semantic mask generation, we utilized the Segment Anything Model (SAM) \cite{kirillov2023segment}, prompted by dermatologists, to accelerate the process. These SAM-generated masks were then subjected to multiple rounds of validation and refinement by dermatological experts.

\textbf{Our contribution}: Firstly, we introduce eSkinHealth, a novel dataset of clinically valuable images collected on-site in West Africa, specifically targeting skin Neglected Tropical Diseases (NTDs) among underrepresented populations to aid in combating these conditions. Secondly, we propose and implement an evidence-based AI-expert collaboration paradigm for efficient multimodal data annotation, leveraging foundation models under expert guidance. Thirdly, we thoroughly analyze the generated annotations and benchmark eSkinHealth across various tasks and modalities, including image-based classification and concept bottleneck models (CBMs), to demonstrate its broad utility and potential to catalyze the development of more equitable and effective diagnostic AI tools.

\begin{figure*}
    \centering
    \includegraphics[width=0.91\linewidth]{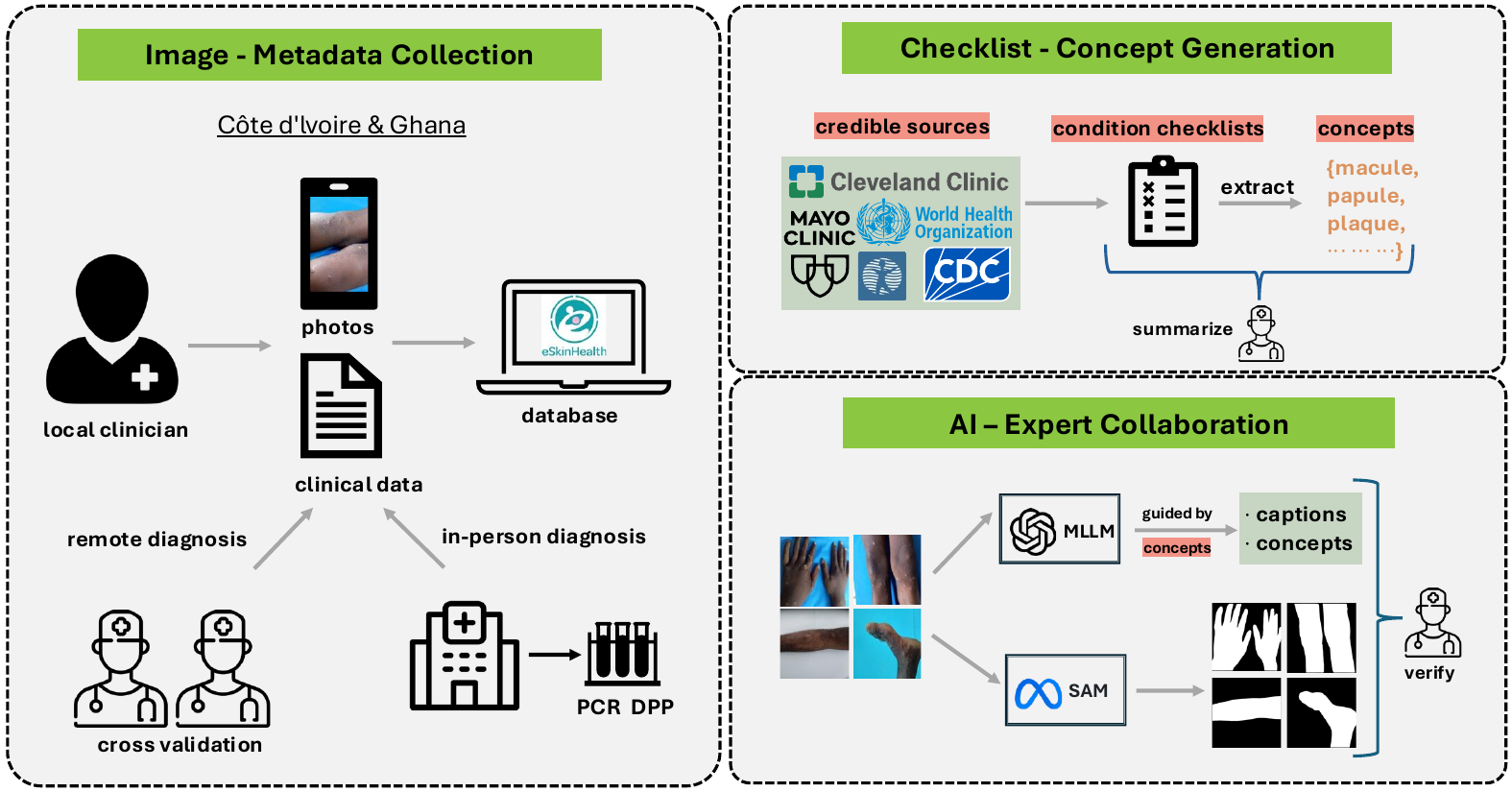}
    \caption{Workflow for eSkinHealth image dataset development. The pipeline consists of three main components: (1) \textbf{Image and Metadata Collection} in Côte d’Ivoire and Ghana, where trained local clinicians capture lesion photos and collect associated clinical metadata, followed by expert diagnosis via remote or in-person review with PCR/DPP testing for select conditions; (2) \textbf{Checklist and Concept Generation}, where condition-specific clinical checklists are curated from credible sources (e.g., WHO, Mayo Clinic, Cleveland Clinic), and concepts are extracted and standardized to encode the lesion information for each image; (3) \textbf{Dermatologist–Expert Collaboration}, where a multimodal large language model (MLLM) and SAM are guided by the checklists to generate per-instance captions and concepts, are then verified by dermatologists to ensure data quality.}
    \label{fig:enter-label}
\end{figure*}

\begin{figure*}
    \centering
    \includegraphics[width=0.91\linewidth]{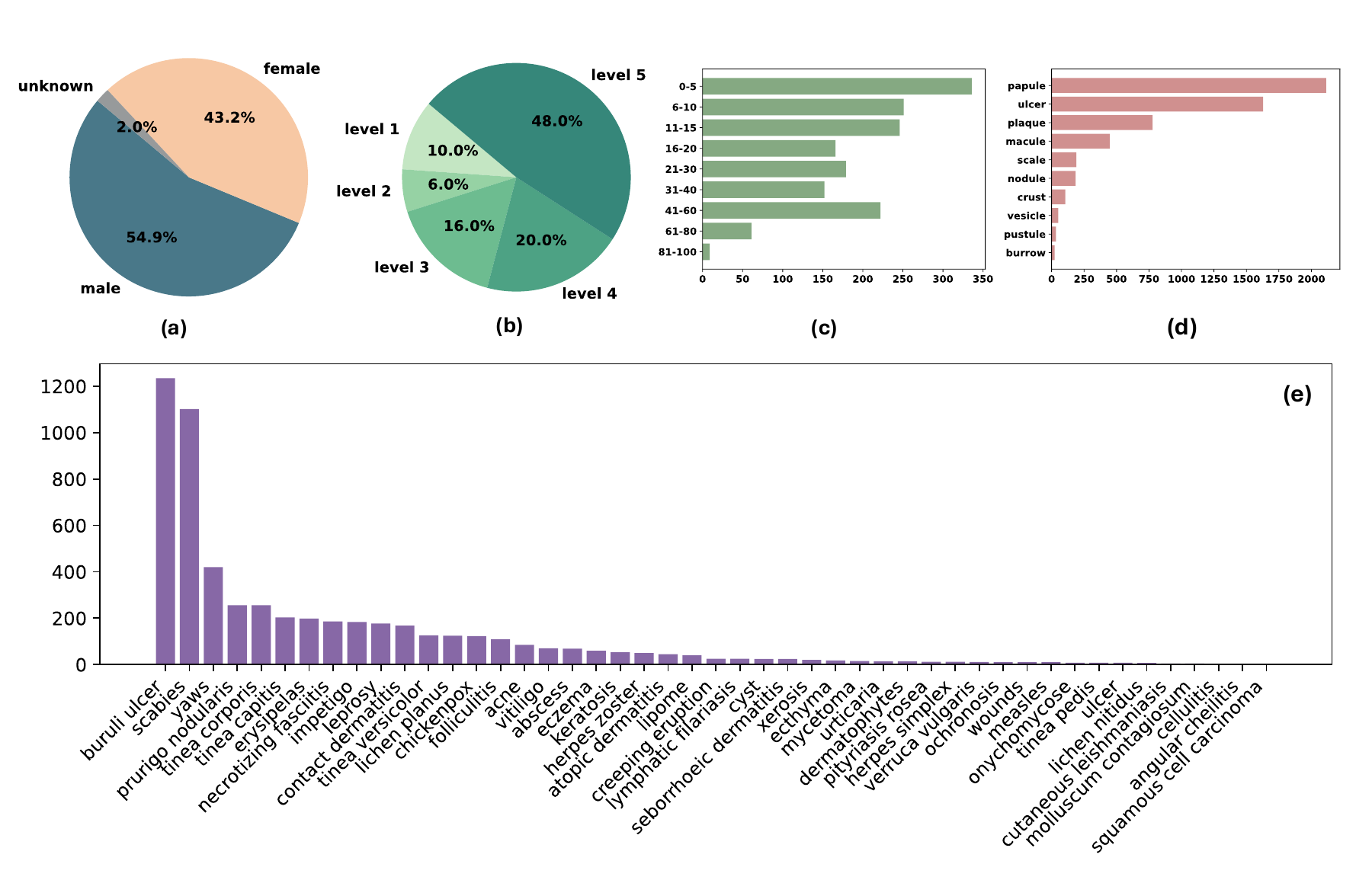}
    \vspace{-2mm}
    \caption{Statistics about the dataset: (a, c) Age group distribution and gender distributions for all cases; (b) clinician rate distribution on the quality of generated caption/concept for sampled instances. Specifically, ``level 5" indicates the highest level of accuracy and ``level 1" indicates the lowest; (d) lesion type distribution for all images; (e) distribution of disease images.}
    \label{fig:distribution}
\end{figure*}

\begin{figure*}
    \centering
    \includegraphics[width=1\linewidth]{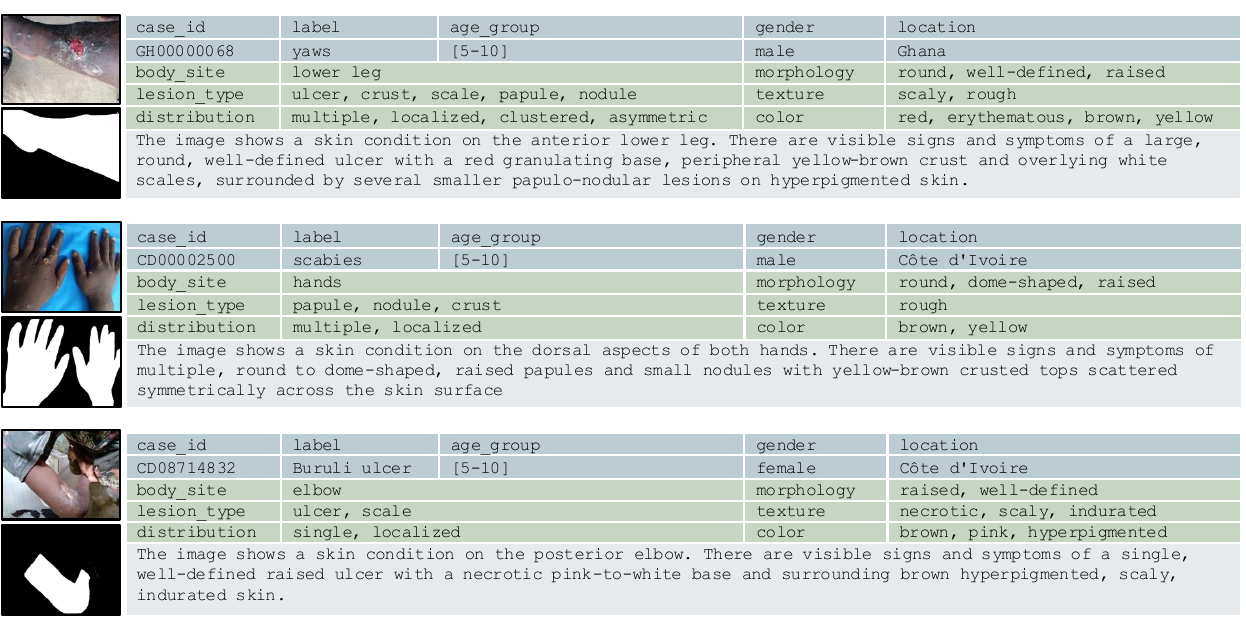}
    \vspace{-3mm}
    \caption{Example of eSkinHealth. For each image, there's a semantic mask generated by SAM and verfied by human. Associated textual data include: (i) clinical metadata (shaded in blue); (ii) instance-level concepts (shaded in green); and (iii) natural language captions (shaded in gray). Concepts and captions were generated by a MLLM and subsequently verified by dermatologists.}
    \vspace{-1.5mm}
    \label{fig:examples}
\end{figure*}

\section{Related Works}
\subsection{Skin Lesion Analysis}
Deep learning models have demonstrated remarkable progress in dermatology, achieving expert-level performance in classifying skin diseases from dermatological images \cite{Esteva2017DermatologistlevelCO, Liu2020-vx, Brinkerarticle, tschandl2020human, soenksen2021using, celebi2019dermoscopy}. Such systems hold significant potential to enhance teledermatology and improve diagnostic accessibility, particularly in resource-limited regions. However, data scarcity still remains a critical bottleneck and hinders the development of robust and reliable predictive models, due to costly and labor-intensive nature of medical data curation and collection. While strategies like synthetic data augmentation and transfer learning have been explored to mitigate this issue, they do not diminish the fundamental need for high-quality, diverse dermatological data \cite{wang2024majority, wang2024achieving, ghorbani2020dermgan, qin2020gan, bissoto2018skin, bissoto2021gan, ali2021enhanced, jain2021deep}. In response, the research community has curated and open-sourced many dermatology datasets, enabling models to generalize better across a wider range of skin conditions, demographic groups, and clinical environments. Table \ref{datasets_comparision} lists the existing datasets in detail. Despite these efforts, gaps remain, particularly concerning the representation of populations affected by skin NTDs and associated disease spectra. To address these gaps, we introduce eSkinHealth, a valuable addition to the community of skin lesion analysis. Our dataset is uniquely compiled from images collected on-site across West Africa—regions where dermatological expertise is limited and AI-driven diagnostic support could make a real impact. 

\subsection{Multimodel Annotation by AI-Expert Collaboration}
Vision-Language Models (VLMs) represent an emerging paradigm for classification tasks, offering the potential to enhance the accuracy and interpretability of skin disease diagnosis by integrating information from multiple modalities \cite{radford2021learning}. Recognizing this potential, recent efforts have focused on creating datasets with richer annotations beyond simple labels. For example, SkinCon \cite{daneshjou2022skincon} and SkinCAP \cite{zhou2024skincap} provide dense clinical concept labels. Additionally, SkinCAP also includes image captions, catering to the growing interest in applying VLMs to dermatology AI. However, generating such detailed multimodal annotations traditionally requires intensive manual effort. SkinCon and SkinCAP, for instance, relied on board-certified dermatologists to manually produce concepts and captions for subsets of existing datasets like Fitzpatrick17k \cite{groh2021evaluating} and DDI \cite{daneshjou2022disparities}. While ensuring high quality, this manual process is resource-intensive and difficult to scale. To address this, human-AI collaboration has emerged as a promising alternative, significantly improving both the efficiency and effectiveness of the annotation process \cite{kim2024meganno+, wang2024human}. This approach synergistically combines the computational power and scalability of AI models, particularly Large Language Models (LLMs), for tasks like generating initial labels or candidate annotations, with the crucial domain-specific knowledge, verification, and refinement provided by human experts. Given that LLMs have demonstrated capabilities in encoding clinical knowledge and acting as collaborators in medical reasoning \cite{singhal2023large, tang2023medagents, zhang2025mllms}, leveraging this AI-expert paradigm for generating multimodal dermatological annotations is a compelling direction, though its application in dermatology remain under-explored.

\begin{table}[h]
\centering
\caption{Image-based classification performance (\%) on 24 largest classes in eSkinHealth. We randomly split the cases into a train and a testing set with a ratio of 0.5.}
\vspace{-3mm}
\small
\resizebox{0.48\textwidth}{!}{%
\begin{tabular}{lccccc}
\Xhline{1pt}
 \textbf{Model} & \textbf{acc} & \textbf{precision} & \textbf{recall} & \textbf{f1} & \textbf{balanced acc} \\ \hline
 ResNet-50      & 54.18     & 33.46     & 30.92     & 31.25     & 30.92                 \\
 ViT-B/16       & 51.70     & 28.40     & 25.85     & 26.03     & 25.85                 \\
 DINOv2         & 51.06     & 27.30     & 25.08     & 25.05     & 25.08                 \\ 
 SwAVDerm       & 53.14     & 29.23     & 25.57     & 26.78     & 25.57                 \\ 
 PanDerm        & \textbf{61.68} & \textbf{42.75} & \textbf{42.11} & \textbf{42.43} & \textbf{42.11}\\ 
                                \Xhline{1pt}
\end{tabular}
}
\label{res: image_classification}
\end{table}

\begin{table}[!ht]
    \centering
    \caption{Classification accuracy (\%) for few-shot settings.}
    \vspace{-3mm}
    \small
    \begin{tabular}{l|cccccc}
        \Xhline{1pt}
        \multirow{2}{*}{\textbf{Method}}& \multicolumn{6}{c}{\textbf{SHOT}} \\ 
         & \textbf{1} & \textbf{2} & \textbf{4} & \textbf{8} & \textbf{16} & \textbf{full} \\
        \midrule
        Linear Probe              & 24.6 & 28.2 & 26.9 & 28.9 & 36.5 & 58.0 \\
        % PCBM            & 6.4   & 13.8  & 15.5   & 23.4   & 28.9   & 51.6   \\
        LaBo            & 26.0 & 27.6 & 28.7 & 25.4 & 33.3 & 55.0 \\
        \Xhline{1pt}
    \end{tabular}
    
    \label{tab:shot_comparison}
\end{table}

\begin{table}[ht]
\centering
\caption{Zero-shot classification performance (\%).}
\vspace{-3mm}
\small
\resizebox{\columnwidth}{!}{%
\begin{tabular}{lccccc}
\Xhline{1pt}
\textbf{Model} & \textbf{acc} & \textbf{precision} & \textbf{recall} & \textbf{f1} & \textbf{balanced acc} \\
\midrule
CLIP   & 18.2 & 9.6  & 17.1 & 13.2 & 13.2 \\
SigLIP & 18.7 & 14.6 & 18.7 & 19.1 & 19.1 \\
\Xhline{1pt}
\end{tabular}
\label{tab:zero_shot}
}
\end{table}

\section{eSkinHealth Benchmark}
\subsection{Image Collection and Labeling}
The images in the eSkinHealth dataset were acquired in Côte d'Ivoire and Ghana as part of an ongoing study utilizing digital health tools for clinical data documentation and teledermatology \cite{yotsu2023mhealth}. Data collection involved capturing photographs of skin lesions alongside comprehensive clinical information, including patient gender, age, lesion body location, past medical and contact history, and detailed disease descriptions (e.g., duration, symptoms like itchiness or pain, progression). This information was gathered to support remote diagnosis by dermatologists. Image acquisition was performed by nurses or community health workers specifically trained in dermatological phototaking by experienced dermatologists. Photographs were taken using the built-in cameras of Lenovo Tab M10 FHD Plus smart tablets under field conditions within rural clinics across six health districts known to be endemic for one or more skin NTDs. The original images, captured at 1920 x 2560 pixel resolution and stored in JPEG format, were subsequently resized by half to facilitate easier data transfer. A rigorous diagnostic process was implemented. For cases assessed remotely, two independent dermatologists (RRY, AD), each with over 10 years of relevant experience in the region, provided diagnoses. Disagreements were resolved through review and discussion with a third dermatologist (BV) to reach a consensus. A subset of cases was diagnosed in person by a dermatologist during project monitoring visits. Furthermore, diagnostic confirmation was sought for specific conditions where appropriate: polymerase chain reaction (PCR) testing was used for a portion of suspected Buruli ulcer cases, and Dual Path Platform (DPP) rapid diagnostic tests (Chembio Diagnostics, Medford, NY, USA) were employed for some suspected yaws cases.

\subsection{Disease Concept and Caption}
\begin{figure}
    \centering
    \includegraphics[width=1\linewidth]{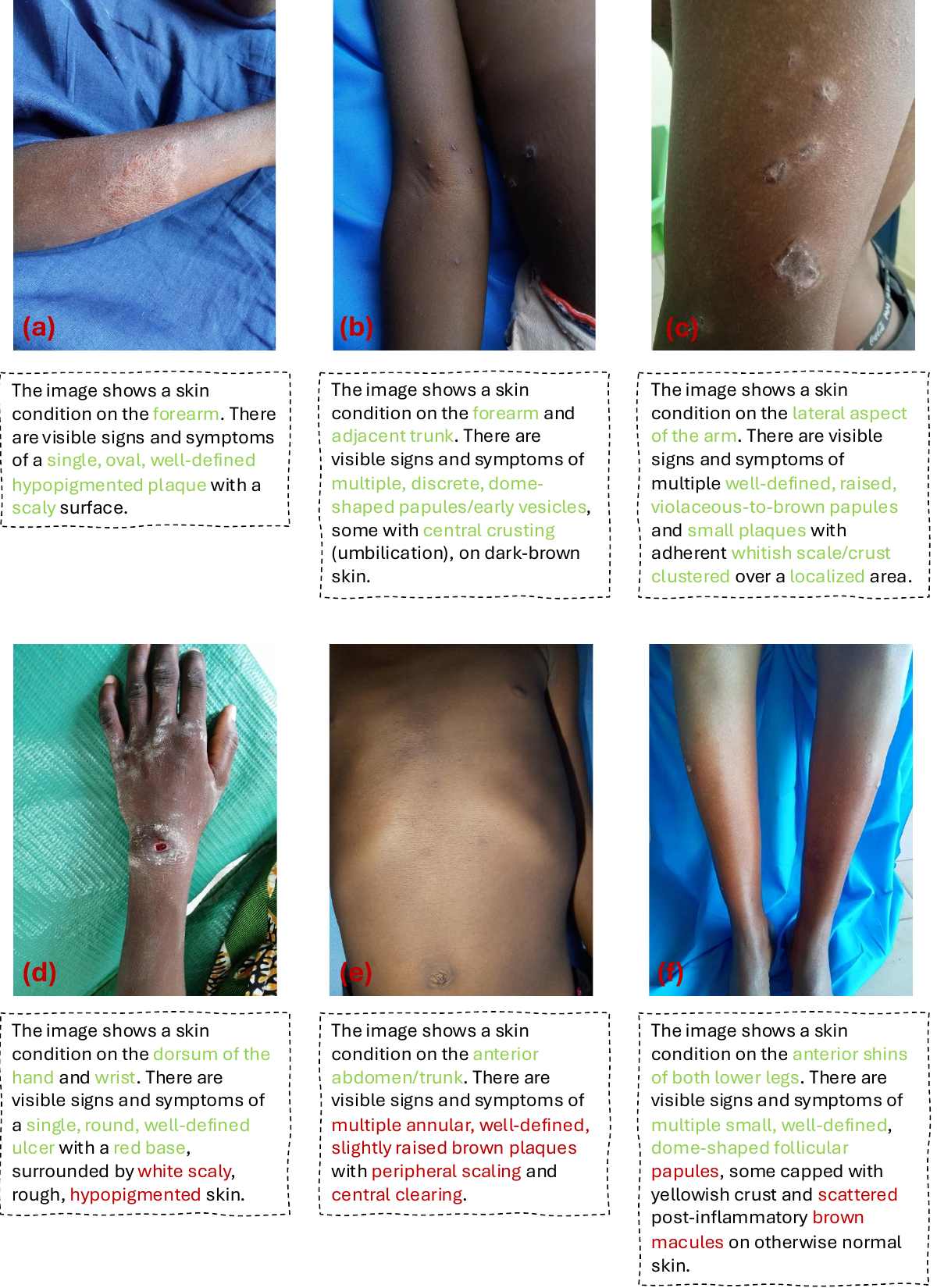}
    \caption{GPT-o1 generated captions on challenging examples highlighted by a dermatologist. (a-c) GPT-o1 accurately described these images. (d-f) Conversely, these cases, challenging even for dermatologists, are not accurate: in (d), GPT-o1 misidentified traditional medicine powder as white, hypopigmented scale; in (e), it failed to recognize small papules (some pustular) and incorrectly implied scaling where none was visible; and in (f), it inaccurately reported "scattered" macules, which instead are few erythematous pustules.}
    \label{fig:clinician_feedback}
\end{figure}

The collection of image captions and concepts involved a multi-stage process. Initially, condition-specific checklists were developed by summarizing clinical information from reputable sources, including DermNet, the World Health Organization (WHO), Mayo Clinic, Cleveland Clinic and CDC. The clinical validity of each checklist was subsequently verified by board-certified dermatologists. Following this verification, we extracted a set of concepts designed to discriminatively encode the clinical features pertinent to each condition. With these verified class-level checklists and organized concepts in place, an advanced Multimodal Large Language Model (MLLM) was utilized to generate instance-specific captions and concepts for every image in the dataset. This generation was guided by the predefined concept vocabulary to ensure consistency and relevance, fostering a collaborative AI-dermatologist framework. In our work, we used OpenAI's GPT-o1 API \cite{jaech2024openai}. To ensure data quality, a randomly sampled 10\% of instances underwent dermatological verification to assess the precision of the generated captions and concepts. As illustrated in Figure $\ref{fig:distribution}$ (b), a dermatologist assigned an accuracy score of 3 or greater to about 84\% of the sampled caption-concept pairs, based on a 5-level scale of image-annotation alignment.  Notably, qualitative observations indicated that for some challenging cases, GPT-o1 provided descriptions comparable to an expert dermatologist. Conversely, in other scenarios that would challenge even a human dermatologist to make a diagnostic decision, GPT-o1 exhibited similar difficulties. Figure \ref{fig:clinician_feedback} presents examples with highlighted comments from a dermatologist illustrating these points. This methodology, where expert-verified concept checklists form the basis for scalable AI-driven annotation, exemplifies a promising collaborative paradigm that significantly reduces processing time and cost while upholding data reliability.

\subsection{Image Segmentation Mask}

To delineate regions of interest (ROIs) crucial for skin diagnosis, we utilized the Segment Anything Model (SAM) to generate masks over the body and lesions \cite{hu2025enhancing}. The process began with clinicians providing multiple initial point prompts (both positive and negative) to guide SAM in identifying the diseased body sites. Subsequently, all generated image-mask pairs underwent manual verification by clinicians. In instances where SAM failed to accurately segment the diseased areas, clinicians refined the segmentation by iteratively adding more positive or negative point prompts for 3 rounds. 

\subsection{Statistics}
In this version of the eSkinHealth dataset, we process 5,623 clinical images from 1,639 distinct cases approved for study. Each case is accompanied by a diagnostic label and patient metadata. The dataset includes 899 male cases and 708 female cases. A notable concentration of cases is observed in younger age groups, particularly those five years old or younger. A visual overview of distributions regarding patient gender, age groups, and disease is provided in Figure \ref{fig:distribution}, with further details available in Appendix \ref{dataset_details}. Furthermore, the dataset features 69 distinct concepts for encoding the clinical information of each image. Correspondingly, we provide discriminative 69-dimensional vectors for each condition based on these concepts. Additionally, instance caption, concepts, and segmentation mask are included for every image in the dataset. More examples of eSkinHealth can be found in Figure \ref{fig:examples}

\section{Experiment}
\vspace{-0.3mm}
\subsection{Experimental Setup}
\vspace{-0.3mm}
To rigorously evaluate the eSkinHealth dataset and provide robust baseline results, we established a consistent experimental setup. All experiments were conducted using defined training and test splits derived from the 1,639 cases, ensuring patient-level separation to prevent data leakage and provide an assessment of generalization.

\vspace{-0.3mm}
\noindent\textbf{Models.} We selected a range of Vision Models (VMs) and Vision-Language Models (VLMs) to benchmark performance on eSkinHealth, aiming to cover diverse architectures and pre-training methodologies. For VMs, we utilized ResNet-50~\cite{resnet}, a widely-adopted convolutional neural network; Vision Transformer (ViT-B/16)~\cite{ViT}, representing prominent transformer-based architectures; and DINOv2~\cite{oquab2023dinov2}, a model pre-trained using advanced self-supervised learning techniques. Additionally, we employed SwAVDerm \cite{shen2024optimizing} and PanDerm \cite{yan2024general}, the most recent self-supervised learning models that were pre-trained on dermatological data. Implementation details are available in Appendix \ref{implementation_details}. 
For VLMs, we employed Contrastive Language-Image Pre-training (CLIP)~\cite{CLIP} and Sigmoid Loss for Language Image Pre-training (SigLIP)~\cite{zhai2023sigmoid} to evaluate zero-shot classification capabilities and the utility of our dataset's image captions. Furthermore, we utilized LaBO~\cite{yang2023language} for concept-driven classification, directly leveraging our expert-curated clinical concepts to enhance interpretability.

\vspace{-0.3mm}
\noindent\textbf{Results.} Classification results for VMs are available in Table \ref{res: image_classification}. PanDerm, a multimodal dermatology foundation model pre-trained through self-supervised learning on over 2 million skin disease images, achieved the best performance. However, the overall classification accuracy is still low, indicating the challenging nature of classifying clinical photos of NTDs, even with state-of-the-art models. Similar observations are made for the few-shot task using a CBM and the zero-shot task using off-the-shelf VLMs, as shown in Table \ref{tab:shot_comparison} and Table \ref{tab:zero_shot} respectively. In these settings, classifying NTDs with no or few training samples is extremely challenging, which aligns with the task's inherent complexity.

\section{Discussion \& Conclusion}
In this work we introduced eSkinHealth, a novel dermatological dataset collected in West Africa, specifically designed to address critical gaps in existing resources for AI development concerning skin Neglected Tropical Diseases (NTDs) and underrepresented populations. We detailed its rigorous collection process involving trained local health personnel and expert dermatological diagnosis, resulting in 5,623 images from 1,639 cases covering 47 distinct skin diseases. A key contribution of our work is the development and implementation of an evidence-based AI-expert collaboration paradigm for comprehensive data annotation. By synergizing the capabilities of advanced foundation models (MLLMs and SAM) with the crucial oversight and validation of board-certified dermatologists, we efficiently generated a rich set of multimodal annotations, including patient metadata, semantic masks, and instance-level captions and concepts, with robust quality control measures. Our extensive baseline experiments have demonstrated the significant utility and versatility of eSkinHealth. The eSkinHealth dataset and our proposed AI-expert collaborative annotation framework represent a significant step towards democratizing AI in dermatology. We believe this resource will be invaluable for researchers and developers working to create more equitable, accurate, and interpretable AI-driven diagnostic tools. Future work will focus on expanding the dataset, exploring more sophisticated modeling techniques tailored to its multimodal nature, and investigating pathways for the clinical validation and deployment of AI tools developed using eSkinHealth to support healthcare workers in endemic regions. Ultimately, we aim for this work to contribute meaningfully to the global effort to combat skin NTDs and improve dermatological care for all.

% \begin{acks}
% \end{acks}

\clearpage

\bibliographystyle{ACM-Reference-Format}
\bibliography{main}

\clearpage
\appendix
\onecolumn

\begin{center}
    \huge \textbf{Appendix}
\end{center}

\section{Dataset Details}
\label{dataset_details}

Table \ref{fig:detailed_disease_distribution} lists the image counts for each disease class.

\begin{table}[]
\centering
\caption{Detailed disease distribution in eSkinHealth.}
\label{fig:detailed_disease_distribution}
\begin{tabular}{lll}
\rowcolor[HTML]{C8E0D9} 
\textbf{condition}      & \textbf{image\_count} & \textbf{case\_count} \\ \hline
scabies                 & 1,127                 & 296                  \\
buruli ulcer            & 1,250                 & 266                  \\
yaws                    & 420                   & 170                  \\
prurigo                 & 258                   & 85                   \\
tinea capitis           & 192                   & 85                   \\
tinea corporis          & 244                   & 70                   \\
erysipelas              & 206                   & 66                   \\
leprosy                 & 198                   & 59                   \\
necrotizing fasciitis   & 181                   & 58                   \\
impetigo                & 184                   & 51                   \\
contact dermatitis      & 173                   & 49                   \\
pityriasis versicolor   & 140                   & 46                   \\
folliculitis            & 116                   & 39                   \\
varicelle               & 125                   & 35                   \\
lichen planus           & 128                   & 31                   \\
acne                    & 89                    & 30                   \\
abscess                 & 68                    & 26                   \\
vitiligo                & 71                    & 25                   \\
eczema                  & 62                    & 19                   \\
atopic dermatitis       & 46                    & 18                   \\
keratosis               & 56                    & 16                   \\
herpes zoster           & 50                    & 16                   \\
lipome                  & 42                    & 13                   \\
mycetoma                & 36                    & 12                   \\
lymphatic filariasis    & 31                    & 10                   \\
seborrhoeic dermatitis  & 24                    & 9                    \\
tinea                   & 23                    & 9                    \\
cyst                    & 26                    & 8                    \\
creeping eruption       & 26                    & 7                    \\
xerosis                 & 20                    & 5                    \\
ecthyma                 & 18                    & 5                    \\
urticaria               & 14                    & 5                    \\
verruca vulgaris        & 11                    & 5                    \\
herpes simplex          & 11                    & 5                    \\
pityriasis rosea        & 12                    & 4                    \\
dermatophytes           & 15                    & 3                    \\
measles                 & 11                    & 3                    \\
wounds                  & 10                    & 3                    \\
tinea pedis             & 8                     & 3                    \\
ulcer                   & 8                     & 3                    \\
onychomycose            & 8                     & 2                    \\
lichen nitidus          & 7                     & 2                    \\
squamous cell carcinoma & 2                     & 2                    \\
ochronosis              & 10                    & 1                    \\
cutaneous leishmaniasis & 4                     & 1                    \\
molluscum contagiosum   & 4                     & 1                    \\
angular cheilitis       & 2                     & 1                    \\
cellulitis              & 2                     & 1                    \\ \hline
\textbf{Total}          & \textbf{5,769}        & \textbf{1,679}      
\end{tabular}
\end{table}

% \begin{table}[ht]
% \centering
% \caption{Concept vocabulary extracted from validated checklists.}
% \begin{tabularx}{\textwidth}{l X}
% \toprule
% \textbf{Category} & \textbf{Concepts} \\
% \midrule
% Lesion type &
% macule, papule, plaque, nodule, vesicle, pustule, bulla, wheal, crust, scale, ulcer,
% fissure, excoriation, atrophy, burrow, sinus tract, kerion, scutulum, horn,
% warty papilloma, erosion \\[2pt]
% \hline

% Distribution / pattern &
% single, multiple, clustered, localized, generalized, symmetric, asymmetric,
% dermatomal, linear \\[2pt]

% \hline

% Morphology &
% oval, round, annular, concentric, dome‐shaped, scalloped, band, targetoid, reticular,
% geometric, raised, flat, well‐defined, sharp, umbilicated \\[2pt]

% \hline

% Texture &
% indurated, hyperkeratotic, cracked, necrotic, scaly, smooth, rough \\[2pt]

% \hline
% Color &
% black, brown, pink, salmon, orange, red, violaceous, purple, yellow, gray, white,
% blue, erythematous, opaque, transparent, translucent, hypopigmented, hyperpigmented \\
% \bottomrule
% \end{tabularx}
% \label{tab:concept-list}
% \end{table}

\begin{table}[htbp] % Using [htbp] for more flexible float placement
\centering
\caption{Concept vocabulary extracted from validated checklists.}
\label{tab:concept-list} % Label after caption
\begin{tabularx}{\textwidth}{l X}
\toprule
\textbf{Category} & \textbf{Concepts} \\
\midrule
Lesion type &
macule, papule, plaque, nodule, vesicle, pustule, bulla, wheal, crust, scale, ulcer,
fissure, excoriation, atrophy, burrow, sinus tract, kerion, scutulum, horn,
warty papilloma, erosion \\
\midrule 

Distribution / pattern &
single, multiple, clustered, localized, generalized, symmetric, asymmetric,
dermatomal, linear \\
\midrule 

Morphology &
oval, round, annular, concentric, dome‐shaped, scalloped, band, targetoid, reticular,
geometric, raised, flat, well‐defined, sharp, umbilicated \\
\midrule 

Texture &
indurated, hyperkeratotic, cracked, necrotic, scaly, smooth, rough \\
\midrule 

Color &
black, brown, pink, salmon, orange, red, violaceous, purple, yellow, gray, white,
blue, erythematous, opaque, transparent, translucent, hypopigmented, hyperpigmented \\
\bottomrule
\end{tabularx}
\end{table}

\begin{table}[htbp] % [htbp] suggests placement: Here, Top, Bottom, or on a separate Page
\centering 

\caption{Example of Checklist for Scabies.} % Standard caption for table environment
\label{tab:scabies_concepts_wrapped}       % Label for cross-referencing

\vspace{-\baselineskip} % Optional: To reduce space between caption and table, as in your original

% Using a minipage to control the width of the lstlisting environment.
% \linewidth here, under \centering, correctly refers to the available text width.
\begin{minipage}{\linewidth}
% The \tiny command for the table environment should propagate here.
% If lstdefinestyle also specified \tiny in basicstyle, it would be redundant but harmless.
% Our current basicstyle inherits the size.
\begin{lstlisting}[style=conceptlist]
1. (*@\textbf{Location}@*): finger-webs, lateral fingers; palmar/wrist flexures and elbows; axillae, belt-line/waist, buttocks; male genitalia, areolae/nipples; in infants also scalp, face, palms, soles, ankles
2. (*@\textbf{Distribution}@*): linear or S-shaped burrows, often in rows/"track lines"; scattered or clustered pruritic papules; can become generalized in children; symmetric involvement of occluded skin folds; sparing of back in classic scabies; widespread hyperkeratotic crusts in crusted (Norwegian) scabies
3. (*@\textbf{Lesion Type}@*): primary: burrows, pinpoint papules, tiny vesicles or pustules; nodules on genitals/axillae; thick hyperkeratotic plaques in crusted scabies. secondary: excoriations, impetiginised crusts.
4. (*@\textbf{Shape}@*): serpiginous / wavy tunnels (burrows); dome-shaped papules or nodules; crusted plaques with fissures in severe cases
5. (*@\textbf{Border}@*): burrow edges well-defined narrow track; papules discrete and round; nodules sharply demarcated; crusted plaques have irregular overhanging edges
6. (*@\textbf{Elevation}@*): burrow slight linear ridge; papule raised few mm; nodule firm, sometimes deeply seated; crust elevated thick keratotic layer
7. (*@\textbf{Texture}@*): smooth or scaly papules; fine scale over burrow entry ("mite-sign"); thick, brittle scale/crust in Norwegian scabies
8. (*@\textbf{Color}@*): erythematous to skin-colored papules; burrows gray-white or skin-colored; rash may look red, brown or gray on darker skin; nodules red-brown; crusts gray-yellow
9. (*@\textbf{Translucency}@*): burrow may show tiny dark dot (mite) at one end; vesicles contain clear serous fluid; pustules if secondarily infected; crusted plaques solid keratin, opaque
\end{lstlisting}
\end{minipage}
\end{table}

\section{Automated Annotation via MLLMs}
\label{sec:app_mllms}
For each pair of data, we use the following prompt to collect text annotations from GPT-o1:\\

\begin{lstlisting}[caption={Python Prompt for Clinical Image Analysis}, label={lst:clinical_prompt}]
prompt = f'''This is a clinical image of {}.
Complete the following 2 independnet task.

TASK 1 Free-text caption:
Describe the lesion(s) by replacing the brackets <>
in the template using any dermatology terms you find appropriate:
The image shows a skin disease on <specific body part>.
There are visible signs and symptoms of <detailed visual descriptions of the skin>.

TASK 2 Structured checklist: Return all concepts visible in the image from the concept_list.
concept_list = [lesion_type: macule, papule, plaque,
nodule, vesicle, pustule, bulla, wheal, crust,
scale, ulcer, fissure, excoriation, atrophy, burrow,
sinus tract, kerion, scutulum, horn, warty papilloma,
erosion;
distribution/pattern: single, multiple,
clustered, localized, generalized, symmetric,
asymmetric, dermatomal, linear;
morphology:
oval, round, annular, concentric, dome-shaped, scalloped,
band, targetoid, reticular, geometric, raised, flat,
well-defined, sharp, umbilicated;
texture: indurated, hyperkeratotic, cracked,
necrotic, scaly, smooth, rough, hypopigmented;
color: black, brown, pink, salmon,
orange, red, violaceous, purple, yellow, gray, white, blue,
erythematous, opaque, transparent, translucent].

OUTPUT FORMAT {{Task 1: <filled template>,
               Task 2: {{lesion_type: papule, ..., distribution/pattern: ...}}
             }}'''
\end{lstlisting}

\section{Implementation Details}
\label{implementation_details}
For the Vision Models (ResNet-50, ViT-B/16, and DINOv2), we performed linear fine-tuning, where only the final classification layer was trained while keeping the pre-trained backbone frozen. This common evaluation protocol assesses the quality of the pre-trained features when adapted to our specific task. Training was conducted using an Adam optimizer~\cite{adam} with a learning rate of $0.01$ for $200$ epochs on our official training split for each model. For VLMs such as CLIP and SigLIP, we performed linear probing on the image features, following \cite{zhou2022conditional}'s reported hyperparameters for robust adaptation. The LaBO concept-based VLM classification was fine-tuned using our designed clinical concepts and the official hyperparameters provided by the LaBO authors to ensure fair comparison and reproducibility. All experiments were conducted on NVIDIA RTX 6000 Ada GPUs. 

To provide a comprehensive assessment of model performance, particularly given the inherent class imbalance often observed in real-world medical datasets like eSkinHealth (as depicted in Figure \ref{fig:distribution} (e), showing a long-tail distribution of diseases), we report a suite of standard evaluation metrics. These include Accuracy (Acc), Precision, Recall, $F1$-score, and Balanced Accuracy (Balanced Acc). While accuracy provides an overall correctness measure, precision quantifies the exactness of positive predictions, and recall measures the completeness of positive predictions. The $F1$-score offers a harmonic mean of precision and recall, providing a single score that balances both. Critically, Balanced Accuracy, calculated as the average of recall obtained on each class, gives a more informative picture of performance on imbalanced datasets by avoiding inflation by high accuracy on majority classes. These metrics collectively allow for a nuanced understanding of model strengths and weaknesses on our dataset.

\section{Detailed Discussion on Potential Applications}
\label{sec:appendix_applications}
The eSkinHealth dataset opens up numerous avenues for advancing AI in dermatology. Its rich, multimodal annotations, including instance-specific visual captions and clinical concepts, make it highly suitable for developing and fine-tuning image captioning models tailored to dermatological images. 
% This can enable the creation of specialized diagnostic tools similar to MedDAM \cite{xiao2025describe}, which generates detailed, region-specific captions for other medical modalities like radiology. 
Furthermore, the dataset's structure is ideal for Vision-Language Model (VLM) fine-tuning, enabling the adaptation of general VLMs to the specific nuances of skin disease presentation and terminology. 

Beyond fine-tuning, eSkinHealth's unique focus on Neglected Tropical Diseases (NTDs) and West African populations, along with its detailed annotations, provides a valuable resource for dermatological VLM pre-training, potentially leading to models with better equity and performance on underrepresented conditions. 
Moreover, the data collected from different clinics and regions can be used to simulate the domain shifts common in real-world clinical deployments. 

The inclusion of expert-curated clinical concepts also directly supports the development and evaluation of concept bottleneck models (CBMs), facilitating more interpretable diagnostic AI by explicitly incorporating clinically relevant features into the model's decision-making. 

The unique characteristics of the eSkinHealth dataset, including its focus on underrepresented populations, multimodal annotations, and diverse data sources, make it a valuable resource for several advanced areas of machine learning research. We detail three promising applications below.

\subsection{Specialized Medical Image Captioning and Generation}
The inclusion of instance-specific visual captions and semantic lesion masks makes the eSkinHealth dataset highly suitable for developing and fine-tuning image captioning models tailored to dermatology. In clinical practice, diagnostic interpretation often relies on precise descriptions of localized regions rather than a global summary of the entire image. The dataset can be used to train models that generate these fine-grained, region-specific descriptions. Moreover, the rich, expert-verified annotations can serve as a foundation for guiding generative models; for instance, recent work has utilized a similar AI-expert collaboration paradigm to provide feedback for diffusion models, enabling the synthesis of medically accurate images for data augmentation~\cite{wang2025doctor}.

This work can enable the creation of specialized diagnostic tools for dermatology, similar to the \textbf{MedDAM} framework~\cite{xiao2025describe}, which was the first comprehensive model to leverage large vision-language models for region-specific captioning in other medical domains like radiology.

\subsection{Parameter-Efficient Fine-Tuning (PEFT)}
The eSkinHealth dataset provides an ideal test bed for adapting large, pre-trained Vision-Language Models (VLMs) to the specific nuances of skin disease. As vision models continue to grow in scale, fully fine-tuning them for every downstream task has become computationally and storage-intensive \cite{Mai_2025_CVPR, jia2022vpt}. \textbf{Parameter-Efficient Fine-Tuning (PEFT)} has emerged as a critical alternative, where only a small subset of parameters is updated.

\textbf{Visual Prompt Tuning (VPT)} is a prominent PEFT method that introduces a small number of extra trainable parameters into the input space of a model's layers while keeping the large pre-trained backbone frozen\cite{jia2022vpt, zeng2024visual, xiao2025visual}. The choice between VPT and full finetuning can depend on the relationship between the pre-training and downstream tasks. For instance, one study found that VPT is often preferable when there is a significant difference in task objectives or when data distributions are very similar \cite{han2024facing}. However, a more recent and comprehensive unifying study concluded that, when properly tuned, PEFT methods \textbf{consistently outperform full fine-tuning} on specialized, low-shot benchmarks like VTAB-1K \cite{Mai_2025_CVPR}. Given that eSkinHealth represents a specialized domain shift from the general-purpose images used to pre-train most large models, it serves as a perfect use case for developing and benchmarking highly effective and efficient PEFT methods~\cite{jia2022vpt, zeng2024visual, xiao2025visual, han2024facing, Mai_2025_CVPR}.

\subsection{Benchmarking for Test-Time Adaptation (TTA)}
A significant challenge in deploying AI models in real-world clinical settings is their performance degradation when encountering domain shifts, such as variations in camera equipment, lighting conditions, or patient populations between different hospitals. \textbf{Test-Time Adaptation (TTA)} aims to solve this by adapting a pre-trained model to a new, unseen target domain using only unlabeled test data \cite{wang2021tent, niu2023towards, zhang2025dpcore, Zhang_2025_WACV}.

The eSkinHealth dataset is uniquely suited to serve as a benchmark for TTA methods. The images were collected \textbf{on-site in Côte d'Ivoire and Ghana} from \textbf{rural clinics across six different health districts}. This inherent diversity in data sources can be used to simulate the continually changing environments that TTA models are designed to handle. Researchers could use this dataset to evaluate the robustness of advanced TTA methods, such as \textbf{DPCore}~\cite{zhang2025dpcore}, which is designed for Continual Test-Time Adaptation (CTTA) and uses a dynamic prompt coreset to adapt to a continuous stream of changing domains while preserving previously learned knowledge. Other TTA methods that could be benchmarked using this dataset include~\cite{wang2021tent, niu2023towards, Zhang_2025_WACV}.

\section{Limitations}
We acknowledge limitations, including non-standardized image formats (e.g., variable size, angle, lighting ) and some repeated images resulting from field collection. While initial dermatologist verification of a 10\% sample of AI-generated captions and concepts showed promising quality, broader validation across the entire dataset is beneficial for consistent quality assurance. Future work will focus on dataset expansion and exploring more sophisticated multimodal modeling techniques.

\end{document}